\documentclass[UTF8]{IEEEtran}
\usepackage{cite}
\usepackage{amsmath,amssymb,amsfonts}
\usepackage{algorithmic}
\usepackage{graphicx}
\usepackage{textcomp}
\usepackage{float}
\usepackage{algorithm}
\usepackage{algorithmic}
\usepackage{multirow}
\usepackage{booktabs} 
\usepackage{tikz}
\usepackage{calc}
\usepackage{bbding}
\usepackage{pifont}
\usepackage{wasysym}
\usepackage{amssymb}
\usepackage{color}
\usepackage{subfigure}
\usepackage{graphicx}
\usepackage{geometry}
\usepackage{pgfplots}
\usepackage{tikz}
\usepackage[switch]{lineno}

\begin{document}
\title{Flow Guidance Deformable Compensation Network for Video Frame Interpolation}
\author{Pengcheng~Lei, Faming Fang and Guixu Zhang
\thanks{Pengcheng Lei, Faming Fang, and Guixu Zhang are with   the School of Computer Science and Technology, East China Normal University, Shanghai 200062, China (e-mail:pengchenglei1995@163.com; fmfang@cs.ecnu.edu.cn; gxzhang@cs.ecnu.edu.cn).
}
}

\maketitle

\begin{abstract}
	Motion-based video frame interpolation (VFI) methods have made remarkable progress with the development of deep convolutional networks over the past years. While their performance is often jeopardized by the inaccuracy of flow map estimation, especially in the case of large motion and occlusion. In this paper, we propose a flow guidance deformable compensation network (FGDCN) to overcome the drawbacks of existing motion-based methods. FGDCN decomposes the frame sampling process into two steps: a flow step and a deformation step. Specifically, the flow step utilizes a coarse-to-fine flow estimation network to directly estimate the intermediate flows and synthesizes an anchor frame simultaneously. To ensure the accuracy of the estimated flow, a distillation loss and a task-oriented loss are jointly employed in this step. Under the guidance of the flow priors learned in step one, the deformation step designs a pyramid deformable compensation network to compensate for the missing details of the flow step. In addition, a pyramid loss is proposed to supervise the model in both the image and frequency domain. Experimental results show that the proposed algorithm achieves excellent performance on various datasets with fewer parameters.
\end{abstract}
\begin{IEEEkeywords}
	Video frame interpolation, motion estimation, motion compensation, deformable convolution, distillation learning.
\end{IEEEkeywords}

\section{Introduction}
\label{introduction}
Video frame interpolation (VFI) is a classical low-level vision task to increase the frame rate of a video sequence.
The recent years have witnessed the rapid development of VFI empowered by the success of deep neural networks~\cite{voxel,Super2018,EQVI}. However, it remains some unsolved problems due to challenges like occlusions, large motion, and lighting changes.
According to whether or not the optical flow information is incorporated, the existing VFI approaches can be roughly divided into two categories: 1) motion-based methods and 2) motion-free methods.

\begin{figure}[t]
	\centering
	\includegraphics[width=0.8\columnwidth]{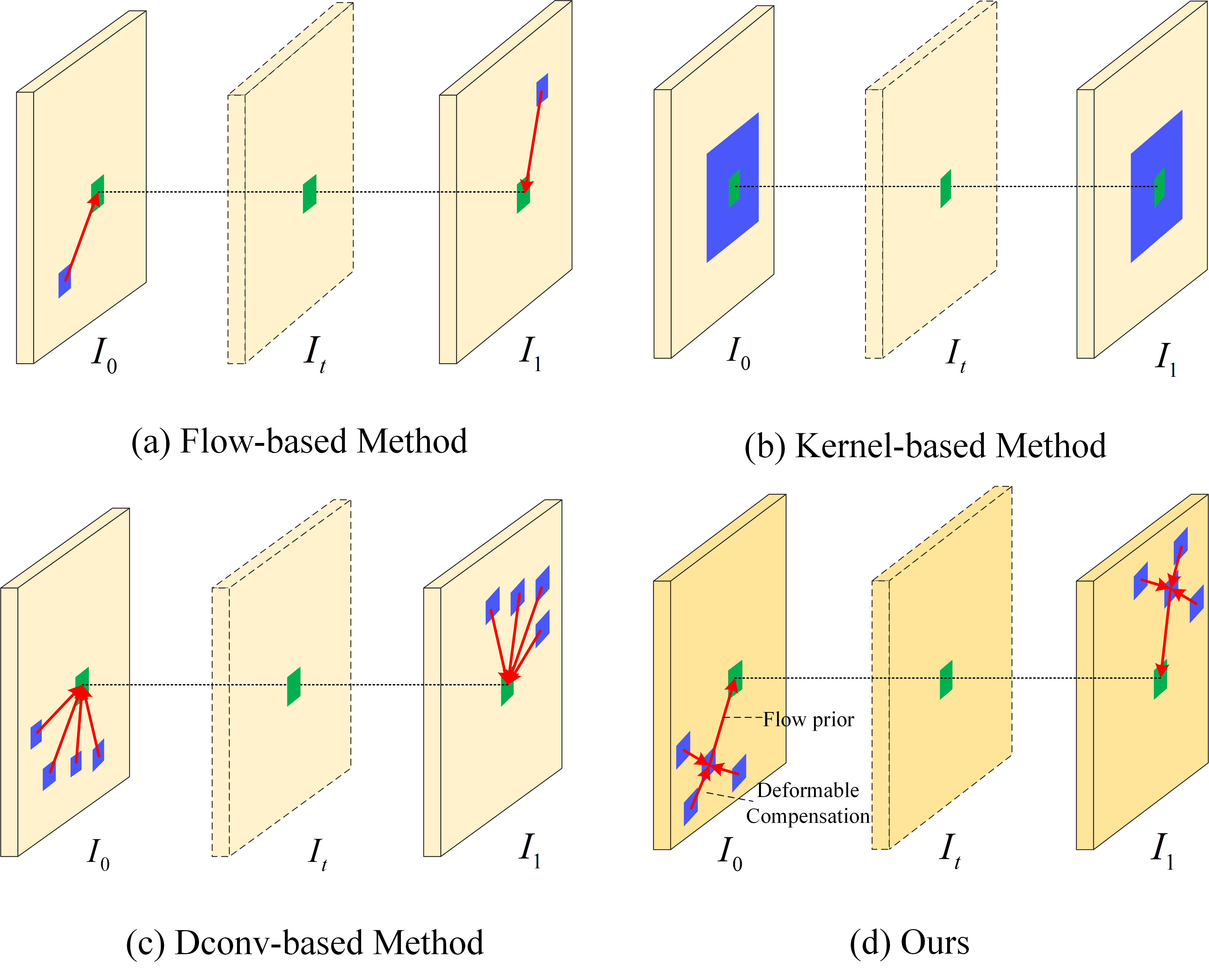} 
	\caption{Overall description of the main streams and our method. (a) Flow-based method, (b) kernel-based method, (c) DConv-based method, (d) our flow guided deformable compensation method. The blue parts of each figure represent the reference points for generating the target pixel.}
	\label{fig:flow}
\end{figure}

In recent years, significant progress has been made by motion-based methods~\cite{Super2018,softmax2020,ABME,IFRNet}. These methods first estimate the bidirectional optical flows and then synthesize the target frame by warping two successive frames forward or backward. In general, these motion-based methods can explicitly represent the motion and reach high fidelity in most cases. However, due to the basic assumptions of the optical flow estimation, e.g. smoothness and consistency, motion-based methods are inherently difficult to handle the interpolation of complicated dynamic scenes which include the regions suffering from occlusion, blur, or abrupt brightness change~\cite{GDConv}.  

Motion-free methods circumvent the need for flow map estimation and consequently are not susceptible to the associated issues. The most representative motion-free methods are the kernel-based methods~\cite{Sep_conv,Adapt_Conv}, which directly generate the target intermediate frame by applying spatially-adaptive convolution kernels to the given frames. Kernel-based methods are effective, however, the rigidity of the kernel shape, severely limits the types of motions that such methods can handle. Indeed, one may need to choose a very large kernel size to ensure enough coverage, which is highly inefficient. 

Most recently, deformable convolution (DConv)~\cite{DCNV1} has been investigated in VFI task to warp features and frames. Gui et al.~\cite{featureflow} propose the Featureflow, which regards DConv as the universal version of motion flows. Different from the one-to-one flow warping, DConv layer produces multiple offsets for each pixel, and uses a weighted average of them to predict target pixel. Thus DConv's offsets warping can be considered as many-to-one flow warping. Theoretically, DConv methods can collect more diverse information and produce more complex transformations, which should be more robust than single optical flow methods when handling the complex motions.
However, in practice, the increased degree of freedom makes the training of the DConv model more difficult, which limits their performance~\cite{PDWN,chan2021understanding}. 

In this paper, we propose a flow guidance deformable compensation network (FGDCN) to overcome the above problems. In Fig.~\ref{fig:flow}, we compare the sampling strategies of different VFI methods. Different from the existing methods, our method decomposes the sampling process into two steps: a flow step and a deformation step. Flow step estimates the intermediate flows to establish the motion correlations between the input frames and the target frame. Since the estimated flow may be inaccurate in some areas, the deformation step further employs the DConv layer to fully explore the abundant local information for compensating the missing details of the single-point flow sampler. The decomposed two-stage learning strategy establishes the connections between the motion-based methods and the DConv-based methods and combines the advantages of both. To be specific, the flow prior makes the DConv layer easier to train, while the DConv layer enhances the transformation ability of the motion-based method and helps it to handle complex motions and recorver more detailed information.

The network structure of the proposed FGDCN also contains a flow step and a deformation step.
The flow step first estimate the intermediate flows and the occlusion mask in a coarse-to-fine mannner and then use them to synthesize an anchor frame. Our coarse flow estimation network is inspired by~\cite{RIFE}, which contains a series of intermediate flow estimation blocks to estimate the intermediate flow directly. Subsequently, a pyramid flow refinement network is employed to gradually refine the coarse intermediate flows. The coarse flows are supervised by the distillation loss and the refined flows are supervised in a task-oriented manner. 
Using the learned accurate flows as guidance, in the deformation step, we first use a pyramid deformable network to compensate for the missing details of the flow step, and then a frame synthesis network is utilized to fuse the pyramid features. To better supervise the pyramid network, we reconstruct the multi-level features simultaneously. This structure is simple but can guide the synthesis network to gradually recover latent sharp images in a coarse-to-fine manner. In addition, a pyramid frequency loss is proposed to constrain the reconstructed images from a global perspective. In summary, our main contributions are as follows:
\begin{itemize}
	\item We propose a novel flow guidance deformable compensation network (FGDCN) for video frame interpolation. To our best knowledge, this is the first model that combines the flow-based with the DConv-based VFI methods in a unified framework. Through our careful design, this model can overcome the shortcomings of a single model and combines the advantages of both.
	\item FGDCN decomposes the interpolation process into two steps: a flow step that employs a coarse-to-fine flow estimation network to estimate the intermediate flows, and a deformation step that uses a pyramid deformable compensation network (PDCN) to compensate for the missing details of the flow step. 
	\item Pyramid reconstruction losses are introduced to supervise the network in both the image and frequency domain. Experimental results show that FGDCN has advantages in handling complex motions and it outperforms the latest state-of-the-art methods with fewer parameters.
\end{itemize}

\begin{figure*}[t]
	\centering
	\includegraphics[width=1.8\columnwidth]{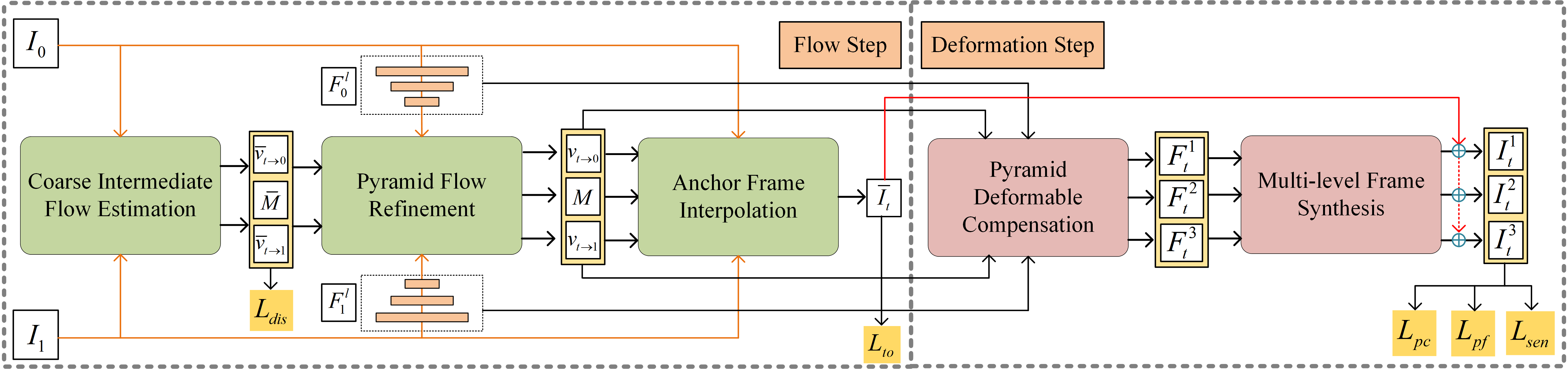} %
	\caption{An overview of the proposed FGDCN. It contains a flow step and deformation step. The flow step consists of a coarse intermediate flow estimation module, a pyramid flow refinement module and an anchor frame interpolation module. The deformation step consists of a pyramid deformable compensation network and a multi-level frame synthesis network.}
	\label{fig:overview}
\end{figure*}

\section{Related Works}
\subsection{Motion-based video frame interpolation}
Video frame interpolation, aiming to synthesize intermediate frames between existing ones of a video, is a longstanding problem. The recent mainstream VFI methods are the motion-based methods, which first estimate the bidirectional optical flows and then use them to warp the input frames. 
Liu et al.~\cite{voxel} introduced a deep network that produced 3D optical flow vectors and warps input frames by trilinear sampling. Recent works have explored a few strategies for improving the performance of such methods. These strategies include utilizing additional contextual information to interpolate high-quality results~\cite{context}, developing unsupervised techniques by cycle consistency~\cite{ICCV2019}, detecting the occlusion by exploring the depth information~\cite{DAIN}, forward warping input frames using softmax splatting~\cite{softmax2020}, leveraging the transformer~\cite{VFIFormer} to model the long-term dependencies, exploring more diverse information using many-to-many flow~\cite{ST-MFNet,M2M}, and constructing efficient architectures for speeding up the interpolation~\cite{IFRNet}. 
Although significant progress has been made by these motion-based VFI approaches, their basic assumption that the motions between neighboring frames are uniform linearity, seriously limits their performance in complex scenarios. 
Recently, a more accurate quadratic model was proposed in~\cite{EQVI} to replace the naive linear model for estimating motion. Nevertheless, the complexities and irregularities of real-world motions cannot be completely captured by a simple mathematical model~\cite{GDConv}. 
In this paper, we employ a pyramid deformable network to compensate for the defects of the motion-based approaches.
\subsection{Intermediate flow estimation}
The intermediate optical flows are highly uncertain, thus how to estimate accurate intermediate optical flows from the consecutive frames is a crucial problem for motion-based VFI methods. When obtaining the intermediate flow, the target image can be obtained by simple backward warping. Jiang et al.~\cite{Super2018} linearly combined optical flows between the input frames to approximate the intermediate flows. Bao et al.~\cite{MEMC} proposed a flow projection layer to reverse the optical flows between the input frames and the target frame. Xue et al.~\cite{Toflow} proposed a task-oriented flow estimation network to predict the intermediate flow directly. Recent VFI methods~\cite{RIFE,IFRNet} proposed to use the distillation strategy to promote the accuracy of the intermediate flow estimation and leads to large performance improvement. Inspired by these works, our method jointly uses the distillation strategy and the task-oriented flow estimation strategy to predict the intermediate flow in a coarse-to-fine manner.

\subsection{Deformable convolution}
Dai et al.~\cite{DCNV1} proposed the deformable convolution network, where additional offsets were calculated to obtain information away from its regular local kernel neighborhood.
Zhu et al.~\cite{DCNV2} further introduced a modulation mask to strengthen the capability of manipulating spatial support regions. 
Let $p_k$ be the $k$-th sampling offset in a standard convolution with kernel size $n\times n$. For example, when $n=3$, we have $p_k\in \{(-1,-1),(-1,0),\cdots,(1,1)\}$. We denote the $k$-th additional learned offset and mask at location $p + p_k$ by $\Delta p_k$ and $\Delta m_k$ respectively. A deformable convolution can be formulated as
\begin{equation}
	F_{out}(p) = \sum^{n^2}_{k=1}w(p_k)\cdot F_{in}(p+p_k+\Delta p_k)\cdot \Delta m_k(p),
\end{equation}
where $F_{in}$ and $F_{out}$ represent the input and output features respectively, $w$ denotes the kernel weights.
Due to its powerful spatial adaption ability, deformable convolution has been widely used in various tasks such as video object detection~\cite{ObjectDetection}, action recognition~\cite{motion}, semantic segmentation~\cite{segmentation} and video super-resolution~\cite{EDVR}. 
For VFI, \cite{Ada_cof} and \cite{EDSC} first introduced the DConv layer in kernel-based methods, where additional offsets were calculated to adaptively sample useful information from the input frames. Gui et al. and Chen et al.~\cite{featureflow,PDWN} used the DConv layer to directly generate the intermediate frame through blending deep features. Shi et al.~\cite{GDConv} proposed a more generalized DConv layer, which extracted the information from the multiple frames to solve the performance-limiting issues of conventional methods. Different from the above methods that directly use the DConv layer to model the complete motions, we employ a flow guidance DConv layer to compensate for the missing details of the motion-based methods.

\subsection{Flow-guided DConv learning}
Some works have proposed to use the flow-guided DConv learning to improve the performance of the video super-resolution (VSR) and video inpainting (VI) tasks. 
Chan et al.~\cite{chan2021understanding} proposed an offset-fidelity loss to guide the offset learning with optical flow. This loss successfully avoids the overflow of offsets and alleviates the instability problem of deformable alignment in VSR task. Subsequently, Chan et al.~\cite{basicvsr++} further proposed a flow-guided alignment layer to align the features from the reference frames to the target frames for VSR task. Li et al.~\cite{li2022towards} designed a flow-guided feature propagation layer for video inpainting (VI) task.

The above methods have demonstrated the benefits of the flow-guided DConv learning on VSR and VI tasks, however, it has not been exploited in the VFI task. As aforementioned, using either motion estimation method or DConv-based method has its inherent problem. Therefore, how to properly combine the flow information with the DConv learning will be very meaningful to solving the problems that VFI faced.

It’s worth noting that models~\cite{basicvsr++,li2022towards} designed for VSR and VI tasks focus on how to align features from the reference frame to the given target frame, while the VFI task focuses on how to synthesize the unknown target frame from its reference frames. The unknown target frame makes the estimated intermediate flows highly uncertain, as well as affects the subsequent DConv learning. To solve this problem, in this paper, we propose to decompose the frame interpolation process into two steps, i.e. flow step and deformation step. The flow step employs a flow estimation network to estimate the intermediate flows directly and synthesize an anchor frame. The deformation step designs a PDCN to compensate for the missing details of the flow step. 
Detailed network structure will be introduced in the next section.
%

\section{Proposed method}
\subsection{Network overview}
Given two input frames $I_0$ and $I_1$, the video frame interpolation is to synthesize an intermediate frame $I_t$. To achieve this goal, we propose FGDCN for accurate frame interpolation. Its overall structure is shown in Fig.~\ref{fig:overview}. It contains a flow step and a deformation step. The flow step first uses a coarse intermediate flow estimation network to predict the coarse intermediate flows $\overline{v}_{t\rightarrow0}$ and $\overline{v}_{t\rightarrow1}$. Then a pyramid network is employed to refine the coarse flows gradually and we get refined flows $v_{t\rightarrow0}$ and $v_{t\rightarrow1}$. The pyramid network contains three levels $l\in\{1, 2, 3\}$. It extracts the pyramid features $F_0^l$ and $F_1^l$ in level $l$ for frame $I_0$ and $I_1$, respectively. At the end of the flow step, we synthesize an anchor frame $\overline{I}_t$ using the learned flows and occlusion mask. The deformation step contains a pyramid deformable compensation network (PDCN) and a multi-level frame synthesis network (MFSN).  PDCN employs the DConv layers to compensate for the missing details of the flow step. MFSN uses the warped pyramid features as input to learn the residuals between the anchor frame and the ground truth. Next, we will introduce each component in detail.
\subsection{Flow step} 
\begin{figure}[h]
	\centerline{\includegraphics[width=8.5cm]{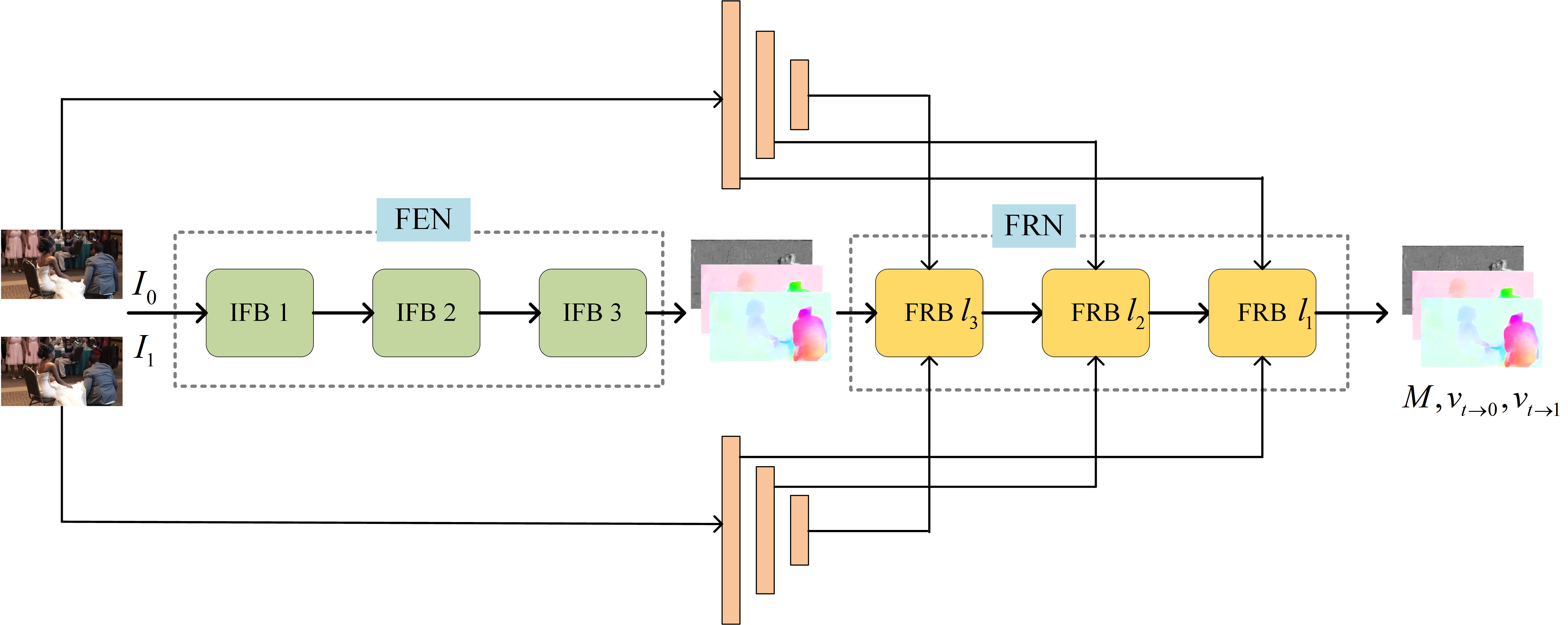}}
	\caption{The structure of the coarse-to-fine flow estimation network.}
	\label{fig:flownet}
\end{figure}

Flow step aims to estimate accurate flow maps and synthesize an anchor frame simultaneously. It contains three parts: a coarse intermediate flow estimation network (FEN), a pyramid flow refinement network (FRN) and an anchor frame interpolation operation. The coarse-to-fine flow estimation network is shown in Fig.~\ref{fig:flownet}. It firstly use a FEN to eastimate the coarse flows and then employ a FRN to refine them. In addition, we also utilize two loss functions, i.e., weakly supervised flow estimation loss and task-oriented flow estimation loss in this step to estimate the intermediate flow from coarse to fine.  
\subsubsection{Coarse intermediate flow estimation} 
We employ FEN to predict the coarse intermediate flow and occlusion mask directly. The network is inspired by~\cite{RIFE}, in which a series of intermediate flow estimation blocks (IFBs) are used to predict the coarse intermediate flows directly. The network is learned by a knowledge distillation scheme. In this paper, we employ the pre-trained flow estimation network Liteflow~\cite{liteflow} as the teacher model to produce the pseudo flow labels. The flow distillation loss can be formulated as:
\begin{equation}
	\label{dis_loss}
	\mathcal{L}_{dis}= ||\overline{v}_{t\rightarrow0}-v_{t\rightarrow0}^{*}||_1 + ||\overline{v}_{t\rightarrow1}-v_{t\rightarrow1}^{*}||_1,
\end{equation}
where $\overline{v}_{t\rightarrow0}$ and $\overline{v}_{t\rightarrow1}$ represents the the estimated intermediate flows, $v_{t\rightarrow0}^{*}$ and $v_{t\rightarrow1}^{*}$ are the pseudo labels. Since the pseudo labels are not entirely accurate, we only use them for weak supervision. To be specific, the weak supervision is conducted in a downsampled resolution instead of the full resolution, which avoids some inaccurate guidance from the pseudo label. Stronger supervision will be given in a task-oriented manner after the subsequent flow refinement network.

\subsubsection{Pyramid flow refinement}
\begin{figure}[t]
	\centering
	\includegraphics[width=0.8\columnwidth]{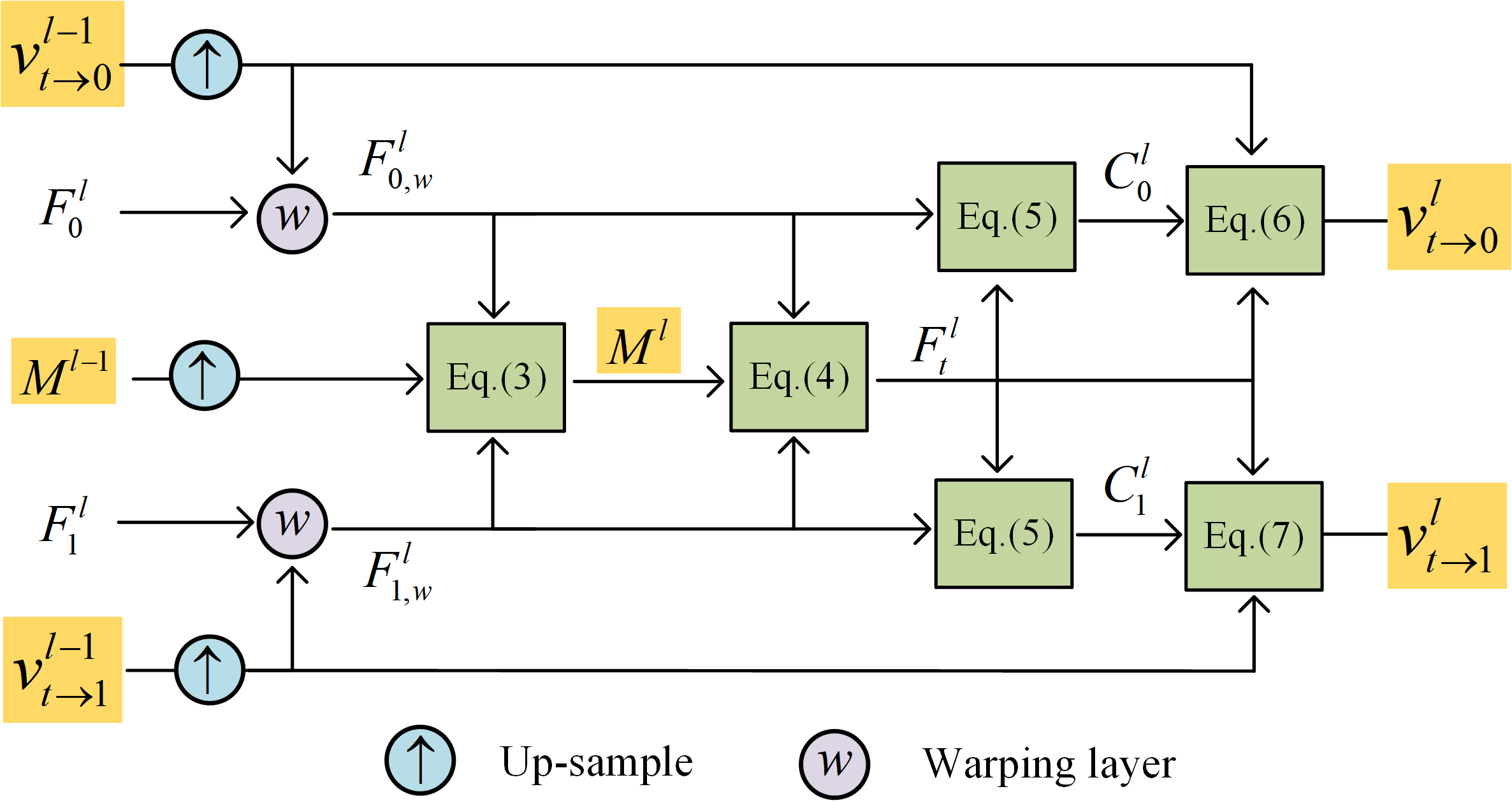} 
	\caption{The structure of the $l$-th flow refinement block (FRB).}
	\label{fig:refine}
\end{figure}

Inspired by~\cite{ABME,VFIFormer}, we employ a pyramid FRN to refine the coarse flows and the occlusion mask.
The structure of the $l$-th flow refinement block is shown in Fig.~\ref{fig:refine}. Firstly, we use the upsampled previous flows to warp the pyramid features and we get the $l$-th warped features $F_{0,w}^l$ and $F_{1,w}^l$. Then the warped features are fed into convolutional layers to update the occlusion mask $M^l$, which is also used to produce a temporary intermediate feature $F_t^l$. These operations can be represented as
\begin{equation}
	M^l = M^{l-1} + \mathcal{T}_{\sigma}(\mathcal{T}_{occ}([F_{0,w}^l, F_{1,w}^l])),
\end{equation}
\begin{equation}
	F_t^l = M^l\odot F_{0,w}^l + (1-M^l) \odot F_{1,w}^l,
\end{equation}
where $M^l$ is the updated occlusion mask in level $l$, $\mathcal{T}_{\sigma}$ and $\mathcal{T}_{occ}$ denotes the sigmoid function and the occlusion network respectively, $\odot$ is the Hadamard product, $[\cdot]$ represents the concatenation operation.

Secondly, we employ a correlation layer~\cite{ABME} to compute matching costs between the warped pyramid feature and the temporary intermediate feature. 
\begin{equation}
	C_0^l, C_1^l = \mathcal{T}_{cor}(F_t^l, F_{0,w}^l, F_{1,w}^l),
\end{equation}
where $\mathcal{T}_{cor}$ denotes the correlation layer, $C_0^l$ and $C_1^l$ are the learned correlations.

Finally, we fuse the correlations with the warped pyramid features using a gated unit to generate the residual of the flow, which is then used to update the input flows.
\begin{equation}
	v_{t\rightarrow0}^{l} = v_{t\rightarrow0}^{l-1} + \mathcal{T}_{gate}([C_0^l, F_{0,w}^l, F_t^l, v_{t\rightarrow0}^{l-1}]),
\end{equation}
\begin{equation}
	v_{t\rightarrow1}^{l} = v_{t\rightarrow1}^{l-1} + \mathcal{T}_{gate}([C_1^l, F_{1,w}^l, F_t^l, v_{t\rightarrow1}^{l-1}]),
\end{equation}
where $\mathcal{T}_{gate}$ repersents the gated unit, $v_{t\rightarrow0}^{l}$ and $v_{t\rightarrow1}^{l}$ are the updated intermediate flows in level $l$.
\subsubsection{Anchor frame interpolation}
We use the refined intermediate flows $v_{t\rightarrow0}$, $v_{t\rightarrow1}$ and the learned occlusion mask $M$ to generate a temporary source frame $\overline{I}_t$, called an anchor frame. The anchor frame can be generated by
\begin{equation}
	\overline{I}_t = M\odot\phi_B(v_{t\rightarrow0}, I_0) + (1-M)\odot\phi_B(v_{t\rightarrow1}, I_1),
\end{equation}
where $\phi_B(\cdot)$ denotes the backward warping function. To estimate an accurate intermediate flow, we train the overall flow estimation network in a task-oriented manner.
\begin{equation}
	\mathcal{L}_{to}= ||I_t^{gt}-\overline{I}_t||_1,
\end{equation}
where $I_t^{gt}$ is the ground truth image. Hence, the full loss function of the flow step is defined as
\begin{equation}
	\mathcal{L}_{fstep}= \mathcal{L}_{to} + \lambda_{dis}\mathcal{L}_{dis},
\end{equation}
where $\lambda_{dis}$ is the loss weight for $\mathcal{L}_{dis}$.

\begin{figure*}[t]
	\centering
	\includegraphics[width=1.8\columnwidth]{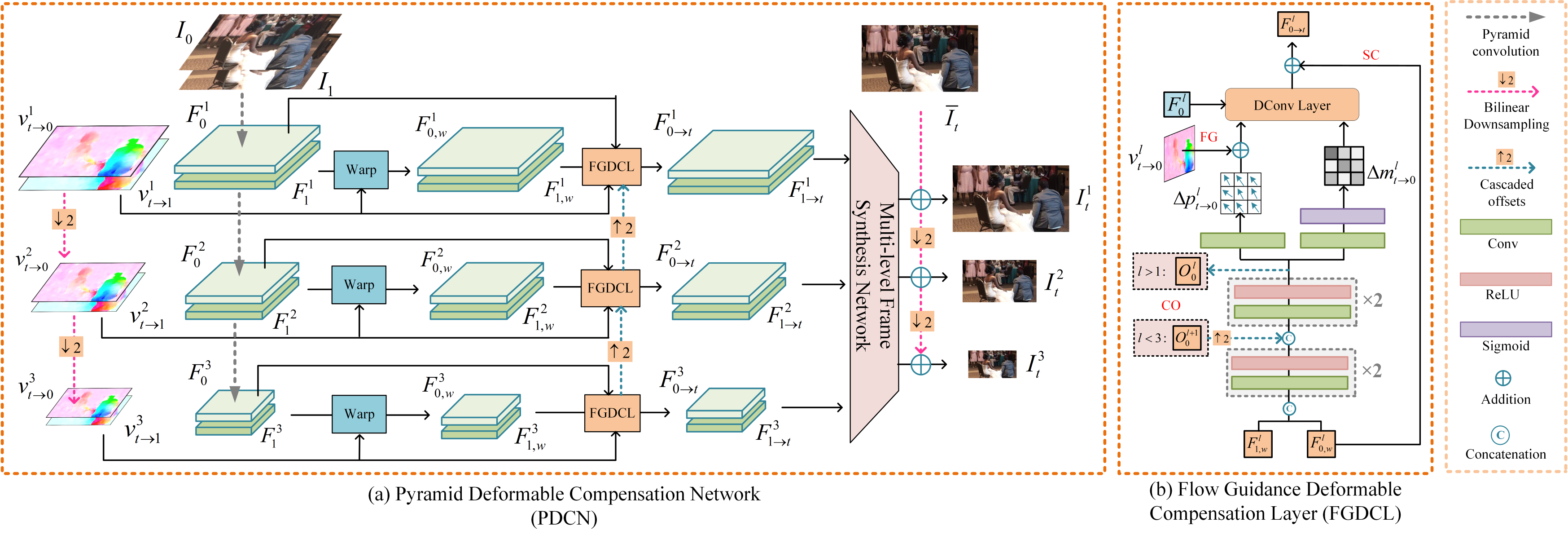} 
	\caption{(a) The framework of the proposed pyramid deformable compensation network (PDCN). (b) The structure of the proposed flow guidance deformable compensation layer (FGDCL).}
	\label{fig:PDCN}
\end{figure*}

\subsection{Deformation step}
As aforementioned, flow-based methods are often jeopardized by the inaccuracy of flow map estimation. Only flow warping on images or features cannot handle the large motions and complex sceneries. To solve this problem, we design a deformation step to compensate for the missing details of the flow step. Its overall structure is shown in Fig.~\ref{fig:PDCN}(a), which contains a pyramid deformable compensation network (PDCN) and a multi-level frame synthesis network (MFSN). Different from the previous DConv-based VFI methods that detect the complete motions directly, our PDCN reduces the burden of the offset learning by using the learned flow priors in the flow step as guidance. This strategy not only makes the DConv layer more trainable but also helps it mine more detailed information.	
\subsubsection{Pyramid deformable compensation network}
PDCN uses a pyramid network and conducts the deformable compensation operations on the multi-level pyramid features. 
To be specific, for the $l$-th level feature $F_0^l$ and $F_1^l$, we first use the learned intermediate flows $v_{t\rightarrow0}^l$ and $v_{t\rightarrow1}^l$ to warp pyramid features and we get the warped feature $F_{0,w}^l$ and $F_{1,w}^l$. Then a flow guidance deformable compensation layer (FGDCL) is employed to compensate for the missing information of the flow warped features.

The architecture of FGDCL is shown in Fig.~\ref{fig:PDCN}(b). It presents the compensation process of the flow warped feature $ F_{0,w}^l$ under the guidance of the flow map $v_{t\rightarrow0}^l$. As shown in the figure, our FGDCL uses the flow warped feature $F_{0,w}^l$ and $F_{1,w}^l$ as inputs and utilizes a series of ``Conv-ReLU" layers to learn the offset residue $\Delta p_{t\rightarrow0}^l$ and the modulation mask $\Delta m_{t\rightarrow0}^l$. Then the offset residues are added by the pre-estimated optical flow and sent to the DConv layer to warp the original feature $F_{0}^l$. Mathematically, the flow guidance DConv layer can be formulated as 
\begin{equation}
	\Delta F_{0,w}^l(p) = \sum^{n^2}_{k=1}w(p_k)\cdot F_{0}^l(p+p_k+v_{t\rightarrow0}^l(p)+\Delta p_k)\cdot \Delta m_k(p),
\end{equation}
where $v_{t\rightarrow0}^l(p)$ denotes the pre-estimated flow offset at position $p$, $\Delta F_{0,w}^l$ is the compensation details for the flow warped feature $F_{0,w}^l$. For convenience, we omit the superscript $l$ and subscript $t\rightarrow0$ of $\Delta p_k$ and $\Delta m_k(p)$. 

At last, we design a skip connection (SC) layer to enforce the DConv layer to learn the residuals based on the flow warped feature. 
\begin{equation}
	F_{0\rightarrow t}^l = \Delta F_{0,w}^l + F_{0,w}^l,
\end{equation}
where $F_{0\rightarrow t}^l$ denotes the final warped feature from the input feature $F_0^l$. Using the same operations, we get the warped feature $F_{1\rightarrow t}^l$ from the input feature $F_1^l$. Both the two warped features are concatenated together to get the pyramid intermediate feature $F_t^l$ and then they are sent to MFSN together to get the final reconstruction results. In addition, we design a cascaded offset (CO) estimation structure in PDCN, which helps the network to estimate more accurate offsets. 
\subsubsection{Multi-level frame synthesis network}
\begin{figure}[t]
	\centerline{\includegraphics[width=7cm]{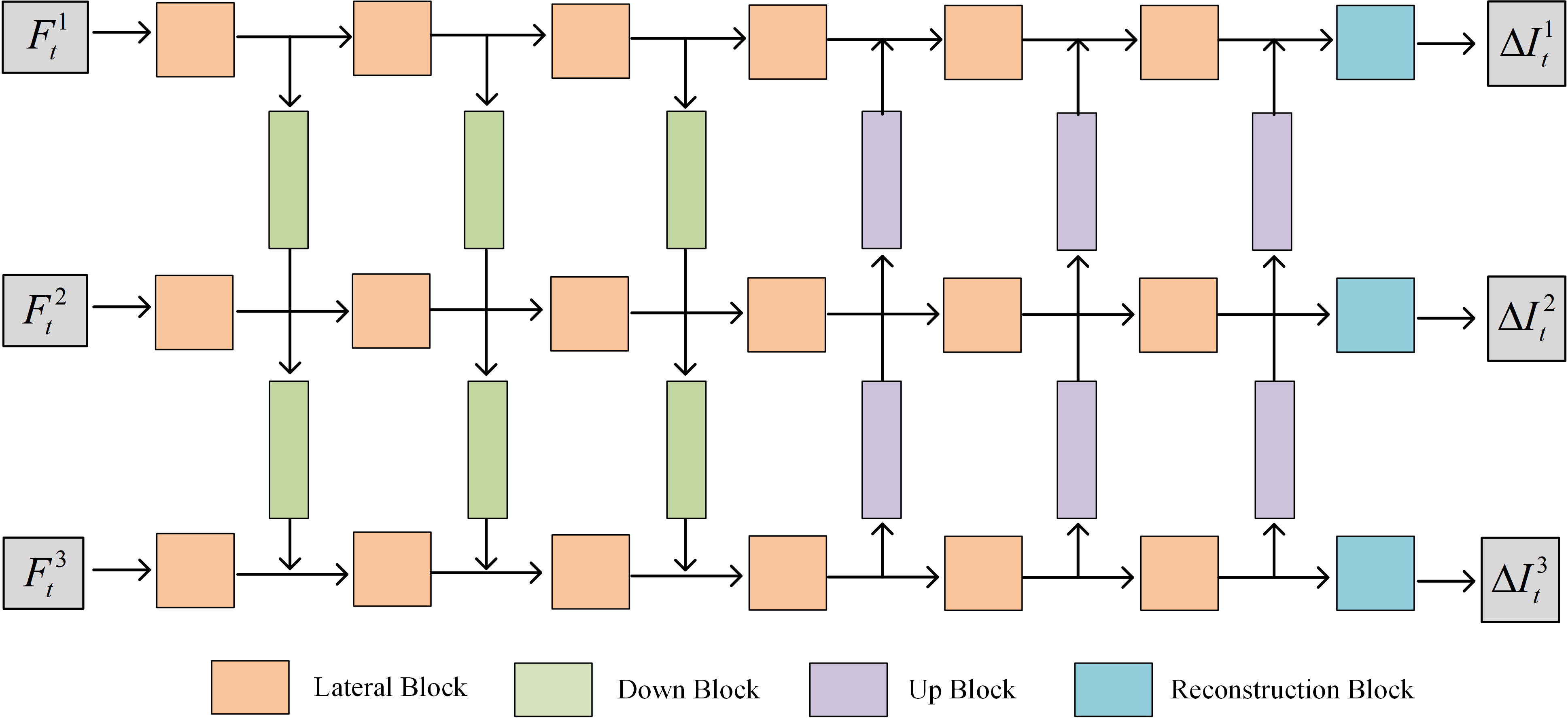}}
	\caption{The structure of the multi-level frame synthesis network (MFSN).}
	\label{fig:gridnet}
\end{figure}

From PDCN, we have got a pyramid of warped features. In this part, we employ MFSN to fuse the multi-level warped features and reconstruct the final results. GridNet~\cite{context} has shown its effectiveness on frame synthesis in various VFI methods~\cite{softmax2020,ABME}. The original version of GridNet has single input and single output, here we design a modified version that has multi-level inputs and reconstruct multi-level outputs simultaneously. Its structure is shown in Fig.~\ref{fig:gridnet}. The network employs the pyramid intermediate features $F_t^l=[F_{0\rightarrow t}^l, F_{1\rightarrow t}^l]$ as inputs and reconstructs the image residuals $\Delta I_t^l$ after the rightmost lateral block. Thus, the final interpolated intermediate frame in the $l$-th level can be obtained by:
\begin{equation}
	I_t^l = \overline{I}_t^l + \Delta I_t^l,
\end{equation}
where $\overline{I}_t^l$ is the anchor frame at the $l$-th level, $\Delta I_t^l$ is the learned image residual between the anchor frame and target frame.

\begin{table*}[t]
	\centering
	\caption{Quantitative results of different VFI methods. The \textbf{bolded} and \underline{underlined} data represent the best and second-best results, respectively.}
	\resizebox{18cm}{2.6cm}{
		\begin{tabular}{lcccccccc}
			\toprule
			\multirow{2}{*}{Method}&\multirow{2}{*}{Params}&\multirow{2}{*}{Vimeo90K$\uparrow$}&\multirow{2}{*}{UCF101$\uparrow$} &\multirow{2}{*}{M.B.$\downarrow$}& \multicolumn{4}{c}{SNU-FILM$\uparrow$}\\
			\cline{6-9}
			&&&&&Easy&Medium&Hard&Extreme\\
			\midrule
			DAIN~\cite{DAIN}&24.0M&34.71/0.9756&34.99/0.9683& 2.04&39.73/0.9902&35.46/0.9780&30.17/0.9335&25.09/0.8584 \\
			CAIN~\cite{CAIN}&42.8M&34.65/0.9730&34.91/0.9690&2.28&39.89/0.9900&35.61/0.9776&29.90/0.9292&24.78/0.8507 \\
			AdaCoF~\cite{Ada_cof}&21.8M&34.47/0.9730& 34.90/0.9680&2.24& 39.80/0.9900&35.05/0.9754&29.46/0.9244&24.31/0.8439 \\
			SoftSplat~\cite{softmax2020}&12.2M&36.10/0.9700&35.39/0.9520&-&-&-&-&- \\
			RIFE~\cite{RIFE}&\underline{9.8M}&35.62/0.9779& 35.28/0.9688&1.96 &40.06/0.9907&35.75/0.9789&30.10/0.9330& 24.84/0.8534 \\
			ABME~\cite{ABME}&18.1M&36.18/0.9805&35.38/0.9698 &2.01&39.59/0.9901&35.77/0.9789&30.58/0.9364&25.42/\textbf{0.8639} \\
			VFIformer-S~\cite{VFIFormer}&17.0M&\underline{36.38/0.9811}&35.36/0.9700&1.90&39.92/0.9905&35.97/0.9793&30.53/0.9366&25.35/\underline{0.8634} \\
			IFRNet-L~\cite{IFRNet} &19.7M&36.20/0.9808&\underline{35.42/0.9698}&\underline{1.89}&40.10/0.9906&\underline{36.12}/0.9797&30.63/0.9368&25.27/0.8609\\
			\midrule
			FGDCN-S&\textbf{6.5M}&36.24/0.9806&\underline{35.42/0.9698}&1.94&\underline{40.31/0.9909}&\underline{36.12/0.9780}&\underline{30.70/0.9376}&\underline{25.51}/0.8631 \\
			FGDCN-L&14.4M&\textbf{36.46/0.9814}&\textbf{35.43/0.9701}&\textbf{1.86}&\textbf{40.40/0.9912}&\textbf{36.23/0.9803}&\textbf{30.82/0.9382}&\textbf{25.58/0.8639} \\
			\bottomrule
		\end{tabular}
	}
	\label{tabel-cmpall}
\end{table*}

\subsection{Loss functions}
\subsubsection{Pyramid content loss}
Following the previous methods, we adopt the $L_1$ loss to supervise the reconstruction contents. Since we reconstruct the multi-level outputs, our pyramid content loss can be represented as
\begin{equation}
	\mathcal{L}_{pc}= \sum^{L}_{l=1} \frac{1}{L}||I_t^{l,gt}-I_t^l||_1,
\end{equation}
where $L$ is the total level of our pyramid network, $I_t^{l,gt}$ and $I_t^{l}$ is the ground truth image and the reconstruction image in level $l$. 
\subsubsection{Pyramid frequency loss}
Existing VFI methods tend to use the pixel-wise loss, e.g. $L_1$ loss, to supervise the reconstruction content, which lacks the supervision of the global information. Here we employ a frequency loss in the Fourier space to introduce global guidance as opposed to pixel-wise loss due to the nature of the Fourier transform~\cite{FFTSR}. Our frequency loss is also carried out on different levels:
\begin{equation}
	\mathcal{L}_{pf}= \sum^{L}_{l=1} \frac{1}{L}||\mathcal{T}_{fft}(I_t^{l,gt})-\mathcal{T}_{fft}(I_t^l)||_1,
\end{equation}
where $\mathcal{T}_{fft}$ denotes the fast fourier transform (FFT) that transfers image signal to the frequency domain. Since the data in frequency domain is complex data that contains amplitude component and phase component, we calculate the loss of each component separately and add them as the total frequency loss.
\subsubsection{Census loss}
Census loss~\cite{UnFlow} $L_{sen}$ is robust to illumination changes, which is widely used in VFI tasks. It is represented as the soft Hamming distance between census-transformed image patches of $I_t^{gt}$ and $I_t$.
\subsubsection{Total loss}
According to the above analysis, our full loss function is defined as
\begin{equation}
	\mathcal{L}_{total}= \mathcal{L}_{pc} + \lambda_{pf}\mathcal{L}_{pf} + \lambda_{dis}\mathcal{L}_{dis} +
	\lambda_{sen}\mathcal{L}_{sen},
\end{equation}
where $\lambda_{pf}$ and $\lambda_{sen}$ are loss weights for $\mathcal{L}_{pf}$ and $\mathcal{L}_{sen}$ respectively. $\mathcal{L}_{dis}$ denotes the distillation loss in Eq.~(\ref{dis_loss}).

\begin{figure*}[h]
	\centering
	\includegraphics[width=1.8\columnwidth]{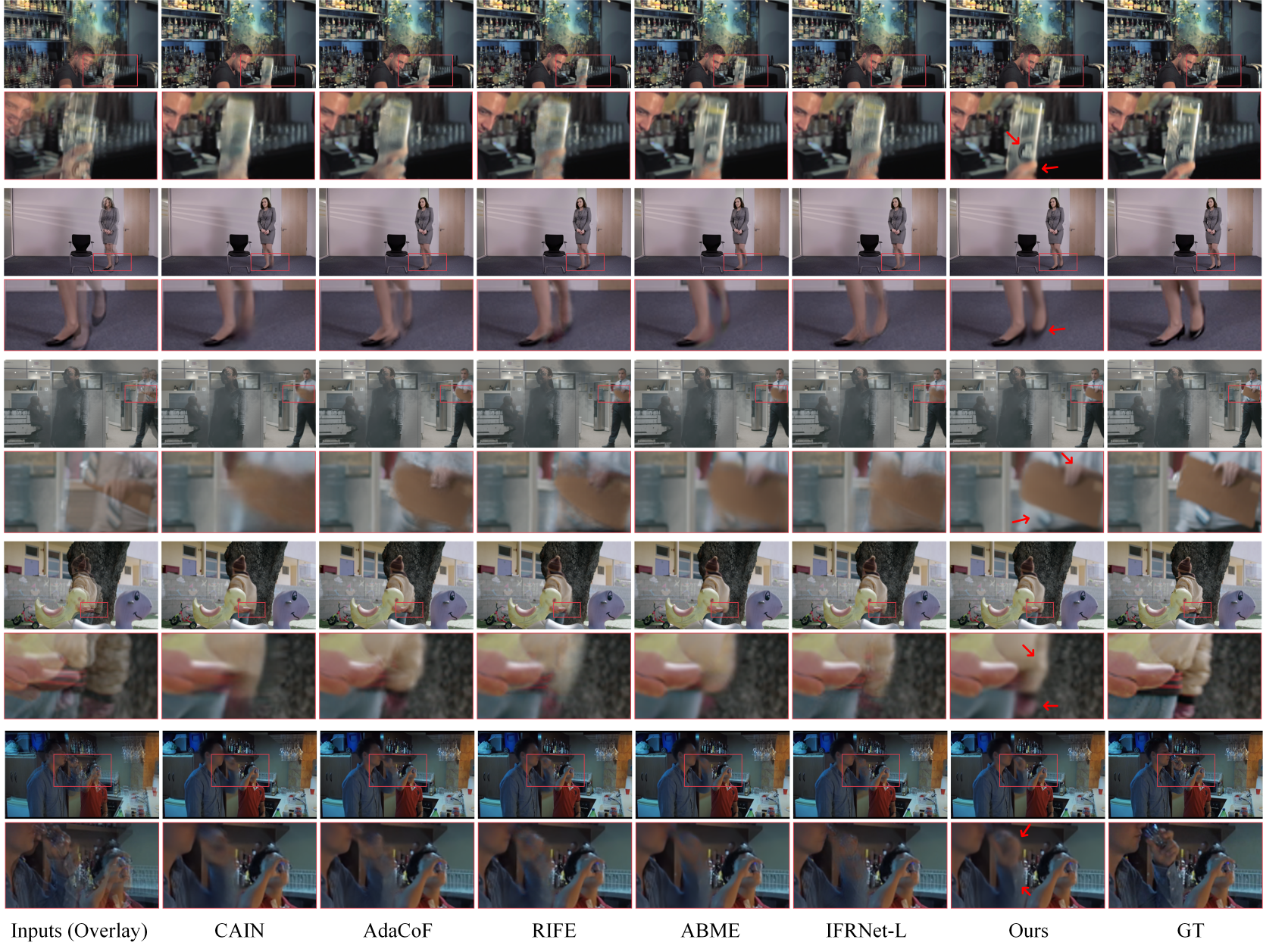} 
	\caption{Visual comparison among different VFI methods on the Vimeo90K testing set.}
	\label{fig:cmp}
\end{figure*}

\begin{figure}
	\centering
	\includegraphics[width=1\columnwidth]{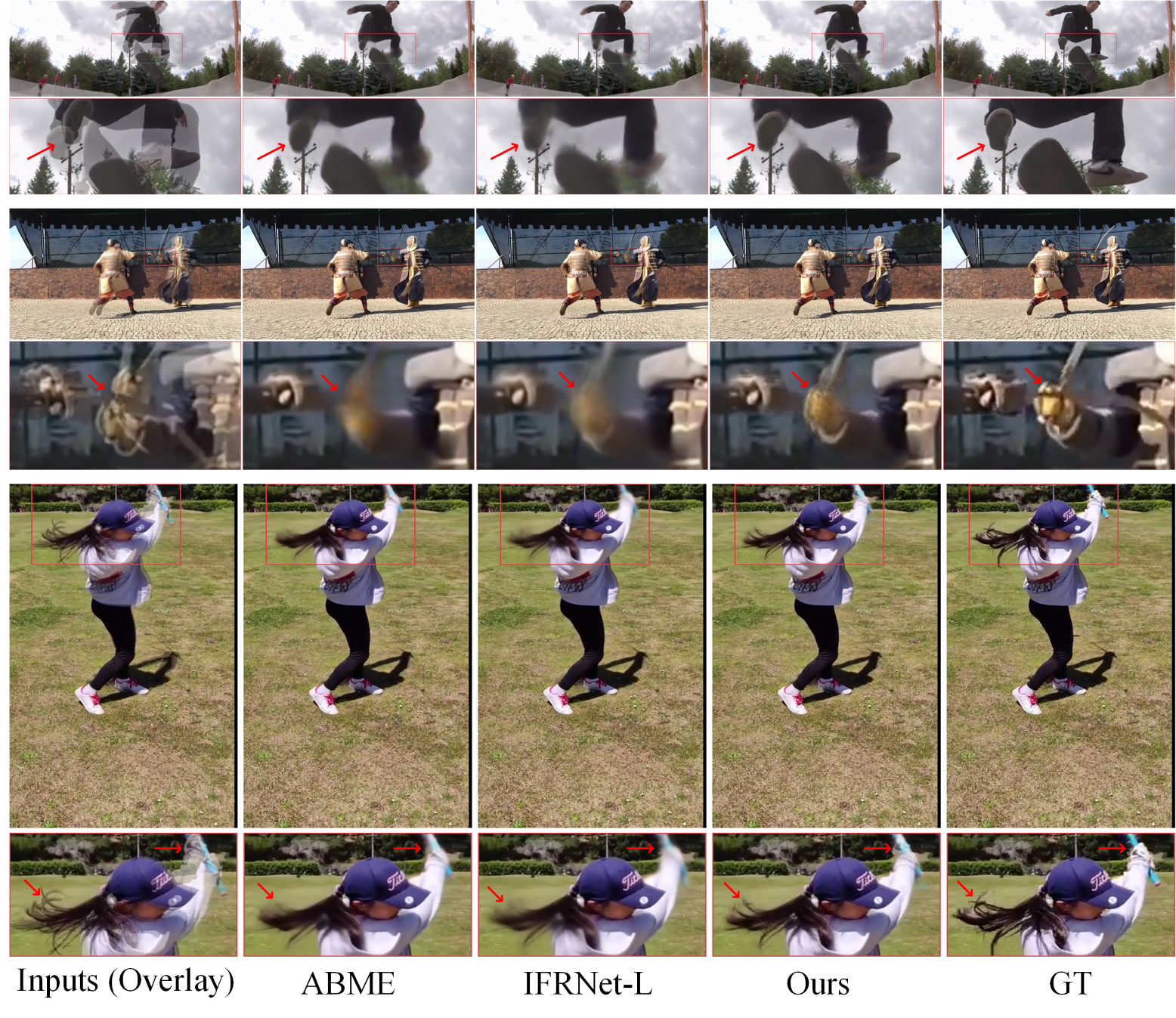} 
	\caption{Visual comparison among existing SOTA VFI methods on the SNU-film testing set.}
	\label{fig:cmp_film}
\end{figure}

\begin{table}
	\centering
	\caption{Running time comparison of recent SOTA methods.}
	\resizebox{9cm}{0.6cm}{
	\begin{tabular}{c|ccc|cc}
		\toprule
		Method & ABME &IFRNet-L& VFIformer-S& FGDCN-S & FGDCN-L \\
		\midrule
		$448\times256$&143ms&\underline{56ms}&181ms&\textbf{40ms} &67ms\\
		\bottomrule
	\end{tabular}
}
	\label{tab-time}
\end{table}

\begin{table*}[t]
	\centering
	\caption{Ablation study on different architecture variants.}
	\resizebox{18cm}{1.6cm}{
		\begin{tabular}{l|lcccccccccc}
			\toprule
			\multirow{2}{*}{}&\multirow{2}{*}{Method}&\multicolumn{4}{c}{Architecture}&\multirow{2}{*}{Params}&\multirow{2}{*}{Vimeo90K}& \multicolumn{4}{c}{SNU-FILM}\\
			\cline{3-6}
			\cline{9-12}
			&&FEN&FRN&PDCN&MFSN&&&Easy&Medium&Hard&Extreme\\
			\midrule
			\multirow{2}{*}{F-Step}&Model 1&\checkmark&&&&4.2M&34.05/0.9711&39.53/0.9900&35.16/0.9774&29.31/0.9256&24.01/0.8382 \\
			&Model 2&\checkmark&\checkmark&&&5.7M&35.33/0.9775&39.97/0.9906&35.46/0.9784&29.73/0.9301&24.36/0.8483 \\
			\cline{1-12}
			\multirow{2}{*}{DConv}&Model 3&&&\checkmark&\checkmark&8.7M&35.49/0.9773&40.21/0.9911&35.42/0.9780&29.57/0.9262&24.27/0.8454 \\
			&Model 3+&&&\checkmark&\checkmark&17.6M&35.75/0.9800&40.28/\textbf{0.9947}&35.53/\textbf{0.9815}&29.65/0.9298&24.30/0.8488 \\
			\cline{1-12}
			\multirow{2}{*}{D-Step}&Model
			4&\checkmark&\checkmark&&\checkmark&13.3M&\underline{36.29/0.9811}&\underline{40.30}/0.9911&\underline{36.08}/0.9802&\underline{30.62/0.9376}&\underline{25.39}/\textbf{0.8640} \\
			&Model 5&\checkmark&\checkmark&\checkmark&\checkmark&14.4M&\textbf{36.46/0.9814}&\textbf{40.40}/\underline{0.9912}&\textbf{36.23}/\underline{0.9803}&\textbf{30.82/0.9382}&\textbf{25.58}/\underline{0.8639} \\
			\bottomrule
		\end{tabular}
	}
	\label{table-cmparc}
\end{table*}

\section{Experiments}
\subsection{Implementation Details}
\subsubsection{Training details}
We implement the proposed algorithm in PyTorch and use Vimeo90K~\cite{Toflow} training set to train FGDCN. We use Adam optimizer to optimize the network. The training process contains two stages. In the first stage, we train the flow step using $\mathcal{L}_{fstep}$ for $2\times10^{5}$ iterations with batch size $24$. The learning rate is set to $1\times10^{-4}$. In the second stage, we train the whole model using $\mathcal{L}_{total}$ in an end-to-end manner for $4\times10^{5}$ iterations with batch size $32$. The learning rate is initially set to $1\times10^{-4}$, and decays to $1\times10^{-5}$ using a cosine attenuation schedule. 
We also attempt to train the entire network directly in an end-to-end manner, but it will lead to slight performance degradation (discuss in ablation study).
The weight coefficients $\lambda_{pf}$, $\lambda_{dis}$ and $\lambda_{sen}$ are 0.1, 0.01 and 1, respectively. We randomly crop $128\times128$ patches from the training samples and augment them by random flip and time reversal.
\subsubsection{Model details}
We design two versions of FGDCN, a FGDCN-Large (FGDCN-L) and a FGDCN-Small (FGDCN-S). In FGDCN-L, the channel numbers of the three IFBs are set to 180, 120, and 90. The channel numbers of the three-level pyramid network are set to 96, 48, and 24. The channel numbers of the different level lateral blocks are set to 192, 96, and 48. In FGDCN-S, the channel numbers of the three IFBs are set to 120, 90, and 60. The channel numbers of the three-level pyramid network are set to 96, 48, and 24. The channel numbers of the different level lateral blocks are set to 96, 48, and 24. 
\subsection{Evaluation metrics and datasets}
We evaluate our method on various datasets covering diverse motion scenes. 
\subsubsection{Vimeo90K~\cite{Toflow}} The Vimeo90K training set contains 51,312 triplets, where each triplet consists of three consecutive video frames with resolution $448 \times 256$. Its testing set contains 3,782 triplets and their resolution are also $448 \times 256$.
\subsubsection{UCF101~\cite{ucf101}} It contains videos with a large variety of human actions. There are 379 triplets with a resolution of $256 \times 256$.
\subsubsection{Middlebury} The Middlebury benchmark is a widely used dataset to evaluate optical flow and VFI methods. Image resolution in this dataset is around $640\times480$. we report the average interpolation error (IE) on the OTHER dataset.
\subsubsection{SNU-FILM~\cite{CAIN}} It contains 1,240 triplets of resolutions up to $1280 \times 720$. There are four different settings according to the motion types: Easy, Medium, Hard and Extreme.

PSNR and SSIM are adopted for quantitative evaluation. Following previous methods~\cite{VFIFormer}, we report the average interpolation error (IE) on Middlebury testing set. A higher PSNR and SSIM, and lower IE indicate better performance.

\subsection{Comparisons with state-of-the-art methods}
We comprehensively compare our FGDCN-S and FGDCN-L with recent state-of-the-art methods, including DAIN~\cite{DAIN}, CAIN~\cite{CAIN}, AdaCof~\cite{Ada_cof}, SoftSplat~\cite{softmax2020}, RIFE~\cite{RIFE}, ABME~\cite{ABME}, VFIformer-Small~\cite{VFIFormer} and IFRNet-Large~\cite{IFRNet}. The quantitative results of these methods are shown in Table~\ref{tabel-cmpall}.\footnote{All the reported results are copied from their original papers.} The bolded and underlined data represent the best and second-best results, respectively. 
It can be observed that our FGDCN-L outperforms recent state-of-the-art methods on all four testing sets with fewer parameters. Our FGDCN-S has the fewest parameters, but it achieves competitive performance compared with recent state-of-the-art methods. We also compare the running times of recent state-of-the-art methods in Table~\ref{tab-time}. The running time is tested on images with $448\times256$ resolution on an NVIDIA RTX3090 GPU. Our FGDCN-L runs $\times3$ faster than VFIformer-S and slightly slower than IFRNet-L. FGDCN-S is the fastest one among these methods. These experiments fully demonstrate the high efficiency of our FGDCN.

Fig.~\ref{fig:cmp} shows the visual comparison of different VFI methods on the Vimeo90K testing set. Fig.~\ref{fig:cmp_film} shows the visual comparison of recent SOTA VFI methods on the SNU-film testing set. All these comparison images contain large and complex motions. Other methods fail to make accurate predictions of these motions, while our method successfully predicts the motions and generates more reasonable results. It indicates that our flow plus DConv compensation learning strategy is capable of handling more complex motions.

\subsection{Ablation studies}
\begin{figure*}[t]
	\centering
	\includegraphics[width=1.8\columnwidth]{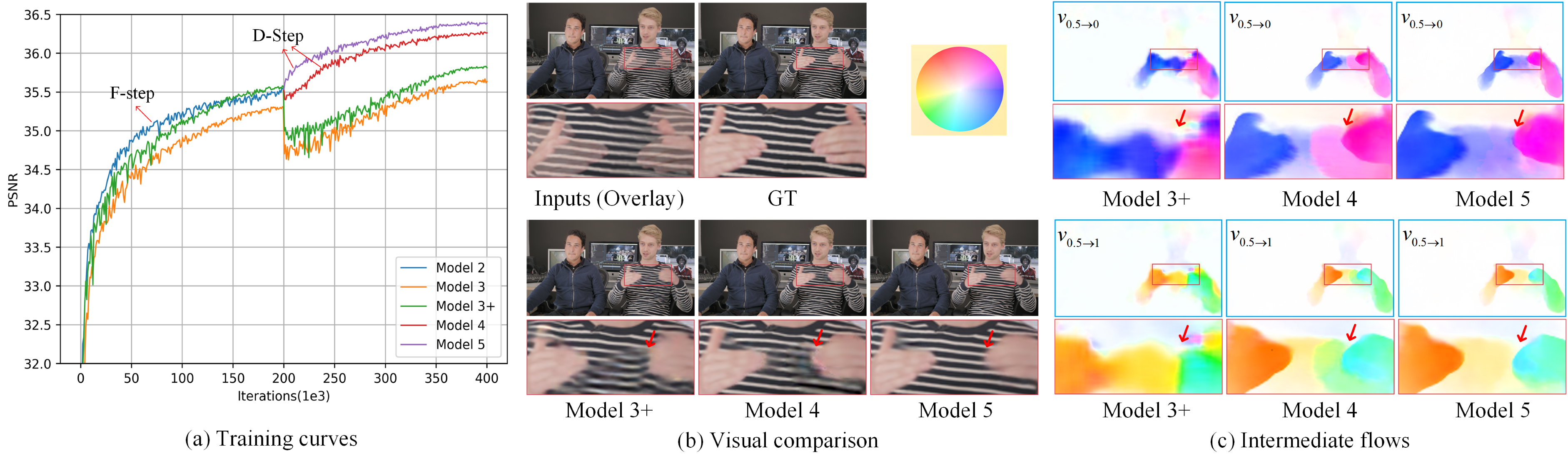} 
	\caption{(a) The training curves of different architecture variants. (b) The reconstruction results on different architecture variants. (c) The learned intermediate flow of different architecture variants.}
	\label{fig:cmp_ab}
\end{figure*}
\begin{figure*}
	\centerline{\includegraphics[width=17cm]{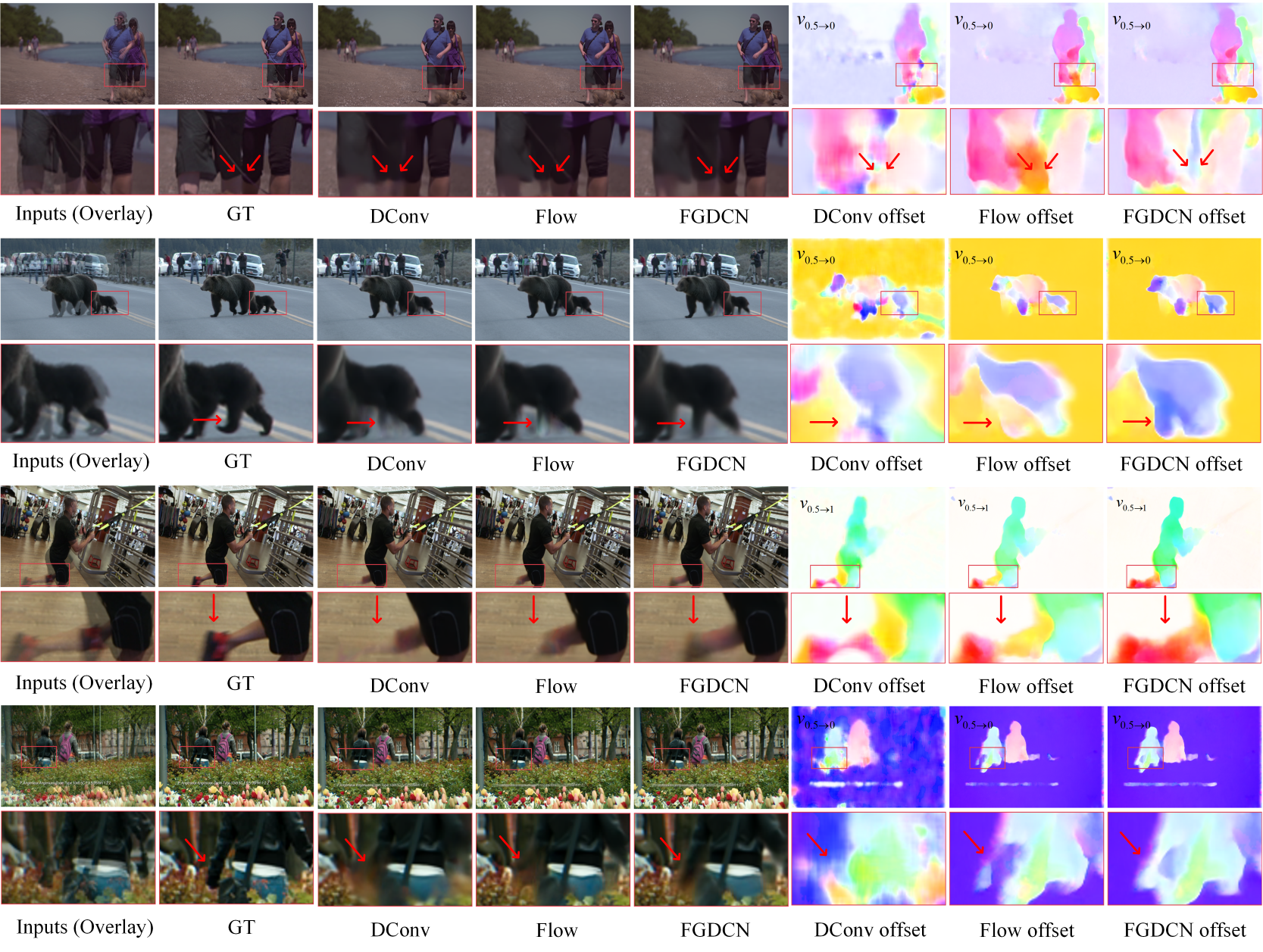}}
	\caption{Visual comparison of the pure DConv-based model, pure flow-based model and our FGDCN model.}
	\label{fig:cmp_offset}
\end{figure*}
\subsubsection{Model architecture ablation} In this part, we decompose our model into four sub-networks, i.e., flow estimation network (FEN), flow refinement network (FRN), pyramid deformable compensation network (PDCN), and multi-level frame synthesis network (MFSN). The performance comparisons are shown in Table~\ref{table-cmparc}. Model 1 and 2 demonstrate the usefulness of the proposed FRN and it improves about 0.5-1dB PSNR on different testing sets. Model 3 is a pure DConv method that uses PDCN to model the complete motions without flow guidance. It works well on simple data, but poorly on complex sceneries. For a fair comparison, we also design an improved version of Model 3, named Model 3+, by increasing its channel numbers. However, the performance gains are very limited. Model 4 is a pure flow-based method. Model 5 introduces PDCN in model 4, its performance improves 0.1-0.25dB on different testing sets with only 1.1M parameters increased. 
Fig.~\ref{fig:cmp_ab} compares the training curves, the reconstruction results, and the learned flows of different models. 
As shown in the figure, model 3+ fails to estimate the accurate motion of hands. Model 4 correctly predicts the motion of hands, but leaves severe artifacts in occluded areas. Model 5 not only captures the accurate motion information but also solves the occlusion problem well. These experiments demonstrate the superiority of the proposed FGDCN compared with the pure flow-based and the pure DConv-based models. More visual comparisons are shown in Fig.~\ref{fig:cmp_offset}.

\subsubsection{Intermediate flow estimation}
\begin{table}[t]
	\centering
	\caption{Ablation study on different intermediate flow estimation methods.}
	\resizebox{8cm}{1.3cm}{
	\begin{tabular}{c|ccc}
		\toprule
		Method & Flow-Net &Params&Vimeo90K\\
		\midrule
		Linear Comb.&PWC-Net&9.4M&34.21/0.9719 \\
		Flow Proj.&PWC-Net&9.4M&32.31/0.9620 \\
		Task-Ori.&Ours&5.7M&34.94/0.9755\\
		Task-Ori.+Dis.&Ours&5.7M&\textbf{35.33/0.9775}\\
		\bottomrule
	\end{tabular}
}
	\label{tab-flow}
\end{table}

\begin{figure}
	\centerline{\includegraphics[width=9cm]{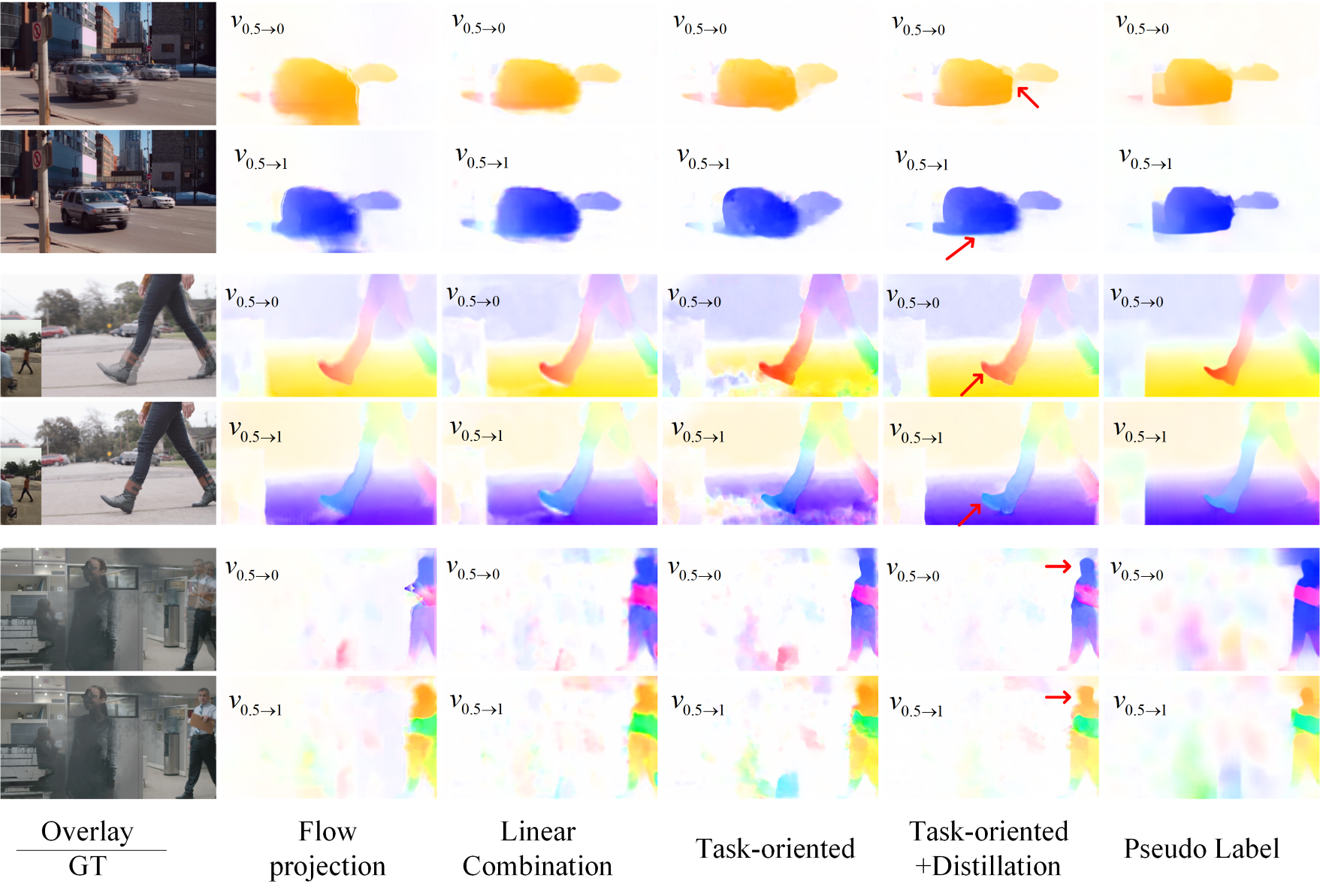}}
	\caption{Visual comparison of different intermediate flow estimation methods on the Vimeo90K testing set.}
	\label{fig:IF_flow}
\end{figure}

Estimating accurate intermediate flows are critical for the VFI task. In our method, reliable intermediate flows can better guide the learning process of the subsequent DConv layer. In this part, we investigate the performance of different intermediate flow estimation strategies. In Table~\ref{tab-flow}, we compare four strategies, i.e., linear combination~\cite{Super2018}, flow projection~\cite{MEMC}, task-oriented~\cite{Toflow} and our distillation plus task-oriented strategy. The first two methods employ PWC-Net~\cite{PWC} as the basic flow estimation network and then predict the intermediate flows using the linear combination and flow projection strategies respectively. The next two methods use our coarse-to-fine flow estimation network to predict the intermediate flow directly. All the models are trained by $2\times10^{5}$ iterations. As listed in the table, the task-oriented strategy is better than the linear combination and flow projection strategies. Besides, the distillation strategy can further improve performance significantly.

We also visually compare the four intermediate flow estimation strategies in Fig.~\ref{fig:IF_flow}. As shown in the figure, the flow projection strategy and the linear combination strategy can approximately estimate the intermediate flows, but the results have severe blur and artifacts at the edge of the objects. The task-oriented strategy directly estimates the intermediate flows with sharp edges. However, due to the lack of supervision from motion priors, the flows may be inaccurate in some cases. Our joint distillation and task-oriented optimization strategy estimate the most accurate intermediate flows.

\subsubsection{FGDCL ablation}
In this paper, we propose a flow guidance deformable compensation layer (FGDCL) to compensate for the missing details of the flow warping. To balance the model performance and complexity, the kernel size and the group number of the DConv layer are set to 3 and 8 in our final model. It means that there are $3\times3\times8=72$ additional offsets and masks will be learned for each position. Here we conduct ablation experiments for the FGDCL. As shown in Fig.~\ref{fig:PDCN}(b), we discuss the effectiveness of the following three components: flow guidance (FG), flow warped skip connection (SC), and cascade offset (CO). Especially, when we remove the flow guidance layer, the original feature $F_0^{l}$ will be replaced by the warped feature $F_{0,w}^{l}$. For fast evaluation, we only train each model for $2\times10^5$ iterations. The quantitative comparisons are shown in Table~\ref{tab-flow}, which indicates the effectiveness of each component in FGDCL.

\begin{figure*}
	\centerline{\includegraphics[width=17cm]{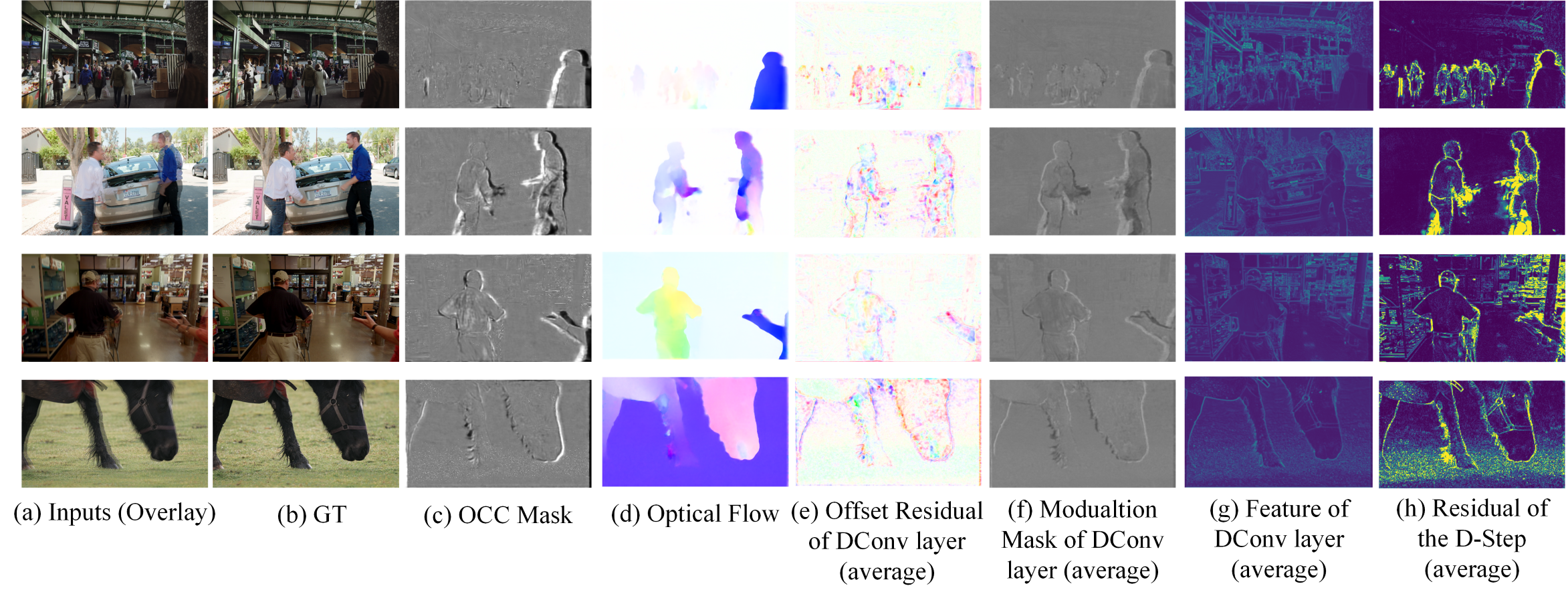}}
	\caption{Visualization of the intermediate component of FGDCN. From left to right, are the input overlaid image, ground truth image, learned occlusion mask $M$, learned optical flow $v_{0.5\rightarrow1}$, learned average offset residual $\Delta p_{0.5\rightarrow0}^{1}$ in the DConv layer, learned modulation mask $\Delta m_{0.5\rightarrow0}^{1}$ in the DConv layer, learned feature $\Delta F_{0,w}^{l}$ in the DConv layer and the learned residual in the deformation step.}
	\label{fig:cmp_flow}
\end{figure*}

In Fig.~\ref{fig:cmp_flow}, we randomly select four testing images to visualize their intermediate components in FGDCN. As shown in the figure, the offset residual focus on the motion areas, especially for the edges of the moving objects, where the optical flows have low fidelity. It indicates that the DConv layer focuses on exploring the missing details of the flow warping. The learned modulation mask of the DConv layer is similar to the occlusion mask. It is more smooth since it is the average value of multiple masks. (g) and (h) represent the learned high-frequency details of the DConv layer and the deformation step, respectively.
\begin{table}[t]
	\centering
	\caption{Ablation study on FGDCL.}
	\resizebox{7.5cm}{1.2cm}{
	\begin{tabular}{ccc|cc}
		\toprule
		FG & SC &CO&Vimeo90K&SNU-hard\\
		\midrule
		&\checkmark&\checkmark&36.12/0.9806&30.58/0.9375 \\
		\checkmark&&\checkmark&36.16/0.9807&30.58/0.9374 \\
		\checkmark&\checkmark&& 36.15/0.9807&30.59/0.9375\\
		\checkmark&\checkmark&\checkmark& \textbf{36.20/0.9809}&\textbf{30.65/0.9376}\\
		\bottomrule
	\end{tabular}
}
	\label{tab-flow}
\end{table}

\begin{table}
	\centering
	\caption{Two-stage training vs. one-stage training.}
	\resizebox{8cm}{1cm}{
		\begin{tabular}{c|cc}
			\toprule
			&Vimeo90K&SNU-hard\\
			\midrule
			One-stage training&36.31/0.9808&30.68/0.9376 \\
			Two-stage training&\textbf{36.46/0.9814}&\textbf{30.82/0.9382} \\
			\bottomrule
		\end{tabular}
	}
	\label{tab-twostage}
\end{table}

\subsubsection{Two-stage training vs. one-stage training}
In this paper, we use a two-stage training strategy to train FGDCN. In the first stage, we train the flow step using $\mathcal{L}_{fstep}$ for $2\times10^{5}$ iterations. In the second stage, we train the whole model using $\mathcal{L}_{total}$ in an end-to-end manner for $4\times10^{5}$ iterations. In this part, we compare the two-stage training strategy with the one-stage training strategy. For the one-stage training, we train the entire network using $\mathcal{L}_{total}$ in an end-to-end manner for $6\times10^{5}$ iterations. As listed in Table~\ref{tab-twostage}, two-stage training outperforms one-stage training by 0.15dB PSNR on the Vimeo90K testing set. We analyze that the pre-training of the flow step provides accurate optical flow for the DConv layer, which can better guide the learning of the DConv layer and enforce it to handle hard examples.

\subsubsection{Loss function ablation}
In this paper, we propose to jointly reconstruct the multi-level images to better optimize the pyramid synthesis network. 
Table~\ref{tab-loss} compares the quantitative results under different combinations of the proposed loss functions. 
Followed by~\cite{VFIFormer}, we regard the single-level content loss $\mathcal{L}_1$, the distillation loss $\mathcal{L}_{dis}$ and the sensus loss $\mathcal{L}_{sen}$ as a baseline. As listed in Table~\ref{tab-loss}, it can be seen that our multi-level content loss $\mathcal{L}_{pc}$ improves PSNR by 0.04dB on Vimeo90K compared with the single-level content loss. When introducing the multi-level frequency loss $\mathcal{L}_{pf}$, the PSNR further improves by 0.07dB on Vimeo90K. It demonstrates that the proposed pyramid image and frequency losses can effectively improve the performance of our model.
\begin{table}
	\centering
	\caption{Ablation study on different loss functions.}
	\resizebox{8cm}{1.1cm}{
	\begin{tabular}{ccc|cc}
		\toprule
		$\mathcal{L}_{baseline}$&$\mathcal{L}_{pc}$&$\mathcal{L}_{pf}$ & Vimeo90K& SNU-hard\\
		\midrule
		\checkmark&&&36.35/0.9812&30.71/0.9381 \\
		\checkmark&\checkmark&&36.39/0.9812&30.76/0.9380 \\
		\checkmark&\checkmark&\checkmark&\textbf{36.46/0.9814}&\textbf{30.82/0.9382}\\
		\bottomrule
	\end{tabular}
}
	\label{tab-loss}
\end{table}

\section{Conclusion}
In this paper, we propose a flow guidance deformable compensation network (FGDCN) for video frame interpolation. To the best of our knowledge, this is the first work that combine the motion-based with the DConv-based methods for solving the VFI problem. To be specific, our method decomposes the frame sampling process into two steps, a flow step that is used to estimate the intermediate flow in a coarse-to-fine manner, and a deformation step that is used to compensate for the missing details of the flow step. This structure successfully establishes the relations between the two methods and combines the strengths of both. Besides, pyramid reconstruction losses are employed to supervise the model in both the image and frequency domain. Both quantitative and qualitative results demonstrate the advantage of our method over the existing state-of-the-art methods.

\bibliographystyle{IEEEtran}     
\bibliography{Reference}

\begin{thebibliography}{10}
\providecommand{\url}[1]{#1}
\csname url@samestyle\endcsname
\providecommand{\newblock}{\relax}
\providecommand{\bibinfo}[2]{#2}
\providecommand{\BIBentrySTDinterwordspacing}{\spaceskip=0pt\relax}
\providecommand{\BIBentryALTinterwordstretchfactor}{4}
\providecommand{\BIBentryALTinterwordspacing}{\spaceskip=\fontdimen2\font plus
\BIBentryALTinterwordstretchfactor\fontdimen3\font minus
  \fontdimen4\font\relax}
\providecommand{\BIBforeignlanguage}[2]{{%
\expandafter\ifx\csname l@#1\endcsname\relax
\typeout{** WARNING: IEEEtran.bst: No hyphenation pattern has been}%
\typeout{** loaded for the language `#1'. Using the pattern for}%
\typeout{** the default language instead.}%
\else
\language=\csname l@#1\endcsname
\fi
#2}}
\providecommand{\BIBdecl}{\relax}
\BIBdecl

\bibitem{voxel}
Z.~Liu, R.~A. Yeh, X.~Tang, Y.~Liu, and A.~Agarwala, ``Video frame synthesis
  using deep voxel flow,'' in \emph{2017 IEEE International Conference on
  Computer Vision (ICCV)}, 2017, pp. 4473--4481.

\bibitem{Super2018}
H.~Jiang, D.~Sun, V.~Jampani, M.-H. Yang, E.~Learned-Miller, and J.~Kautz,
  ``Super slomo: High quality estimation of multiple intermediate frames for
  video interpolation,'' in \emph{2018 IEEE/CVF Conference on Computer Vision
  and Pattern Recognition (CVPR)}, 2018, pp. 9000--9008.

\bibitem{EQVI}
Y.~Liu, L.~Xie, L.~Siyao, W.~Sun, Y.~Qiao, and C.~Dong, ``Enhanced quadratic
  video interpolation,'' in \emph{Computer Vision -- ECCV 2020 Workshops},
  2020, pp. 41--56.

\bibitem{softmax2020}
S.~Niklaus and F.~Liu, ``Softmax splatting for video frame interpolation,'' in
  \emph{2020 IEEE/CVF Conference on Computer Vision and Pattern Recognition
  (CVPR)}, 2020, pp. 5436--5445.

\bibitem{ABME}
J.~Park, C.~Lee, and C.-S. Kim, ``Asymmetric bilateral motion estimation for
  video frame interpolation,'' in \emph{Proceedings of the IEEE/CVF
  International Conference on Computer Vision (ICCV)}, October 2021, pp.
  14\,539--14\,548.

\bibitem{IFRNet}
L.~Kong, B.~Jiang, D.~Luo, W.~Chu, X.~Huang, Y.~Tai, C.~Wang, and J.~Yang,
  ``Ifrnet: Intermediate feature refine network for efficient frame
  interpolation,'' in \emph{Proceedings of the IEEE/CVF Conference on Computer
  Vision and Pattern Recognition (CVPR)}, 2022.

\bibitem{GDConv}
Z.~Shi, X.~Liu, K.~Shi, L.~Dai, and J.~Chen, ``Video frame interpolation via
  generalized deformable convolution,'' \emph{IEEE Transactions on Multimedia},
  vol.~24, pp. 426--439, 2022.

\bibitem{Sep_conv}
S.~Niklaus, L.~Mai, and F.~Liu, ``Video frame interpolation via adaptive
  separable convolution,'' in \emph{2017 IEEE International Conference on
  Computer Vision (ICCV)}, 2017, pp. 261--270.

\bibitem{Adapt_Conv}
------, ``Video frame interpolation via adaptive convolution,'' in \emph{2017
  IEEE/CVF Conference on Computer Vision and Pattern Recognition (CVPR)}, 2017,
  pp. 2270--2279.

\bibitem{DCNV1}
J.~Dai, H.~Qi, Y.~Xiong, Y.~Li, G.~Zhang, H.~Hu, and Y.~Wei, ``Deformable
  convolutional networks,'' in \emph{2017 IEEE/CVF International Conference on
  Computer Vision (ICCV)}, 2017, pp. 764--773.

\bibitem{featureflow}
S.~Gui, C.~Wang, Q.~Chen, and D.~Tao, ``Featureflow: Robust video interpolation
  via structure-to-texture generation,'' in \emph{2020 IEEE/CVF Conference on
  Computer Vision and Pattern Recognition (CVPR)}, 2020, pp. 14\,001--14\,010.

\bibitem{PDWN}
Z.~Chen, R.~Wang, H.~Liu, and Y.~Wang, ``Pdwn: Pyramid deformable warping
  network for video interpolation,'' \emph{IEEE Open Journal of Signal
  Processing}, vol.~2, pp. 413--424, 2021.

\bibitem{chan2021understanding}
K.~C. Chan, X.~Wang, K.~Yu, C.~Dong, and C.~C. Loy, ``Understanding deformable
  alignment in video super-resolution,'' in \emph{Proceedings of the AAAI
  conference on artificial intelligence}, vol.~35, no.~2, 2021, pp. 973--981.

\bibitem{RIFE}
\BIBentryALTinterwordspacing
Z.~Huang, T.~Zhang, W.~Heng, B.~Shi, and S.~Zhou, ``{RIFE:} real-time
  intermediate flow estimation for video frame interpolation,'' \emph{CoRR},
  vol. abs/2011.06294, 2021. [Online]. Available:
  \url{https://arxiv.org/abs/2011.06294}
\BIBentrySTDinterwordspacing

\bibitem{context}
S.~Niklaus and F.~Liu, ``Context-aware synthesis for video frame
  interpolation,'' in \emph{2018 IEEE/CVF Conference on Computer Vision and
  Pattern Recognition}, 2018, pp. 1701--1710.

\bibitem{ICCV2019}
F.~Reda, D.~Sun, A.~Dundar, M.~Shoeybi, G.~Liu, K.~Shih, A.~Tao, J.~Kautz, and
  B.~Catanzaro, ``Unsupervised video interpolation using cycle consistency,''
  in \emph{2019 IEEE/CVF International Conference on Computer Vision (ICCV)},
  2019, pp. 892--900.

\bibitem{DAIN}
W.~Bao, W.-S. Lai, C.~Ma, X.~Zhang, Z.~Gao, and M.-H. Yang, ``Depth-aware video
  frame interpolation,'' in \emph{2019 IEEE/CVF Conference on Computer Vision
  and Pattern Recognition (CVPR)}, 2019, pp. 3698--3707.

\bibitem{VFIFormer}
L.~Lu, R.~Wu, H.~Lin, J.~Lu, and J.~Jia, ``Video frame interpolation with
  transformer,'' in \emph{Proceedings of the IEEE/CVF Conference on Computer
  Vision and Pattern Recognition (CVPR)}, June 2022, pp. 3532--3542.

\bibitem{ST-MFNet}
D.~Danier, F.~Zhang, and D.~Bull, ``St-mfnet: A spatio-temporal multi-flow
  network for frame interpolation,'' in \emph{Proceedings of the IEEE/CVF
  Conference on Computer Vision and Pattern Recognition (CVPR)}, June 2022, pp.
  3521--3531.

\bibitem{M2M}
P.~Hu, S.~Niklaus, S.~Sclaroff, and K.~Saenko, ``Many-to-many splatting for
  efficient video frame interpolation,'' in \emph{Proceedings of the IEEE/CVF
  Conference on Computer Vision and Pattern Recognition (CVPR)}, June 2022, pp.
  3553--3562.

\bibitem{MEMC}
W.~Bao, W.-S. Lai, X.~Zhang, Z.~Gao, and M.-H. Yang, ``Memc-net: Motion
  estimation and motion compensation driven neural network for video
  interpolation and enhancement,'' \emph{IEEE Transactions on Pattern Analysis
  and Machine Intelligence}, vol.~43, no.~3, pp. 933--948, 2021.

\bibitem{Toflow}
T.~Xue, B.~Chen, J.~Wu, D.~Wei, and W.~T. Freeman, ``Video enhancement with
  task-oriented flow,'' \emph{International Journal of Computer Vision}, vol.
  127, pp. 1573--1405, 2019.

\bibitem{DCNV2}
X.~Zhu, H.~Hu, S.~Lin, and J.~Dai, ``Deformable convnets v2: More deformable,
  better results,'' in \emph{2019 IEEE/CVF Conference on Computer Vision and
  Pattern Recognition (CVPR)}, 2019, pp. 9300--9308.

\bibitem{ObjectDetection}
G.~Bertasius, L.~Torresani, and J.~Shi, ``Object detection in video with
  spatiotemporal sampling networks,'' in \emph{ECCV}, 2018, pp. 342--357.

\bibitem{motion}
K.-N. Mac, D.~Joshi, R.~Yeh, J.~Xiong, R.~Feris, and M.~Do, ``Learning motion
  in feature space: Locally-consistent deformable convolution networks for
  fine-grained action detection,'' in \emph{2019 IEEE/CVF International
  Conference on Computer Vision (ICCV)}, 2019, pp. 6281--6290.

\bibitem{segmentation}
L.~Deng, M.~Yang, H.~Li, T.~Li, B.~Hu, and C.~Wang, ``Restricted deformable
  convolution-based road scene semantic segmentation using surround view
  cameras,'' \emph{IEEE Intelligent Transportation Systems Society}, vol.~21,
  no.~10, pp. 4350--4362, 2020.

\bibitem{EDVR}
X.~Wang, K.~C. Chan, K.~Yu, C.~Dong, and C.~C. Loy, ``Edvr: Video restoration
  with enhanced deformable convolutional networks,'' in \emph{IEEE/CVF
  Conference on Computer Vision and Pattern Recognition (CVPR) Workshop}, 2019,
  pp. 1954--1963.

\bibitem{Ada_cof}
H.~Lee, T.~Kim, T.-y. Chung, D.~Pak, Y.~Ban, and S.~Lee, ``Adacof: Adaptive
  collaboration of flows for video frame interpolation,'' in \emph{2020
  IEEE/CVF Conference on Computer Vision and Pattern Recognition (CVPR)}, 2020,
  pp. 5315--5324.

\bibitem{EDSC}
X.~Cheng and Z.~Chen, ``Multiple video frame interpolation via enhanced
  deformable separable convolution,'' \emph{IEEE Transactions on Pattern
  Analysis and Machine Intelligence}, pp. 1--1, 2021.

\bibitem{basicvsr++}
K.~C. Chan, S.~Zhou, X.~Xu, and C.~C. Loy, ``Basicvsr++: Improving video
  super-resolution with enhanced propagation and alignment,'' in
  \emph{Proceedings of the IEEE/CVF Conference on Computer Vision and Pattern
  Recognition (CVPR)}, June 2022, pp. 5972--5981.

\bibitem{li2022towards}
Z.~Li, C.-Z. Lu, J.~Qin, C.-L. Guo, and M.-M. Cheng, ``Towards an end-to-end
  framework for flow-guided video inpainting,'' in \emph{Proceedings of the
  IEEE/CVF Conference on Computer Vision and Pattern Recognition}, 2022, pp.
  17\,562--17\,571.

\bibitem{liteflow}
T.-W. Hui, X.~Tang, and C.~C. Loy, ``Liteflownet: A lightweight convolutional
  neural network for optical flow estimation,'' in \emph{2018 IEEE/CVF
  Conference on Computer Vision and Pattern Recognition}, 2018, pp. 8981--8989.

\bibitem{CAIN}
M.~Choi, H.~Kim, B.~Han, N.~Xu, and K.~M. Lee, ``Channel attention is all you
  need for video frame interpolation,'' \emph{Proceedings of the AAAI
  Conference on Artificial Intelligence}, vol.~34, no.~07, pp.
  10\,663--10\,671, Apr. 2020.

\bibitem{FFTSR}
D.~Fuoli, L.~Van~Gool, and R.~Timofte, ``Fourier space losses for efficient
  perceptual image super-resolution,'' in \emph{Proceedings of the IEEE/CVF
  International Conference on Computer Vision (ICCV)}, October 2021, pp.
  2360--2369.

\bibitem{UnFlow}
S.~Meister, J.~Hur, and S.~Roth, ``Unflow: Unsupervised learning of optical
  flow with a bidirectional census loss,'' \emph{Proceedings of the AAAI
  Conference on Artificial Intelligence}, vol.~32, no.~1, Apr. 2018.

\bibitem{ucf101}
K.~Soomro, A.~R. Zamir, and M.~Shah, ``{UCF101:} {A} dataset of 101 human
  actions classes from videos in the wild,'' \emph{CoRR}, vol. abs/1212.0402,
  2012.

\bibitem{PWC}
D.~Sun, X.~Yang, M.-Y. Liu, and J.~Kautz, ``Pwc-net: Cnns for optical flow
  using pyramid, warping, and cost volume,'' in \emph{2018 IEEE/CVF Conference
  on Computer Vision and Pattern Recognition (CVPR)}, 2018, pp. 8934--8943.

\end{thebibliography}

\end{document}